%% file: main.tex
\documentclass[letterpaper, 10 pt, conference]{ieeeconf}  

\IEEEoverridecommandlockouts                              

\usepackage{microtype}

\usepackage{amsmath,amssymb,amsfonts}
\usepackage{algorithmic}
\usepackage{graphicx}
\usepackage{textcomp}
\usepackage{xcolor}
\usepackage{subfig}
\usepackage{gensymb}
\usepackage{caption}
\usepackage{hyperref}
\usepackage{mwe}
\usepackage{amsfonts}
\usepackage{amssymb}
\usepackage{verbatim}
\usepackage{amsbsy}
\usepackage{siunitx}
\usepackage{tabularx, booktabs}
\usepackage{dblfloatfix}
\usepackage{cite}

\title{\LARGE \bf
Deep Reinforcement Learning for Mapless Navigation of a\\Hybrid Aerial Underwater Vehicle with Medium Transition
}

\author{Ricardo B. Grando$^{1}$, Junior C. de Jesus$^{1}$, Victor A. Kich$^{2}$, Alisson H. Kolling$^{2}$, \\ Nicolas P. Bortoluzzi$^{1}$, Pedro M. Pinheiro$^{1}$, Armando A. Neto$^{3}$, Paulo L. J. Drews-Jr$^{1}$
\thanks{$^{1}$Ricardo B. Grando, Junior C. de Jesus, Nicolas P. Bortoluzzi and P. L. J. Drews-Jr are with the NAUTEC, Centro de Ciencias Computacionais, Univ. Fed. do Rio Grande - FURG, RS, Brazil.
E-mail: {\tt\small paulodrews@furg.br}}
\thanks{$^{2}$Victor A. Kich and Alisson H. Kolling are with the Universidade Federal de Santa Maria - UFSM, RS, Brazil.}
\thanks{$^{3}$Armando A. Neto is with the Electronic Engineering Dep., Univ. Fed. de Minas Gerais, MG, Brazil.
E-mail: {\tt\small aaneto@cpdee.ufmg.br}}
}

\begin{document}

\maketitle
\thispagestyle{empty}
\pagestyle{empty}


\input{0_0_abstract.tex}

\input{1_introduction.tex}
\input{2_related_works.tex}
\input{3_methodology.tex}
\input{4_experimental_results.tex}
\input{5_discussion.tex}
\input{6_conclusion.tex}

\input{7_acknowledgment.tex}


\input{8_references.tex}


\end{document}

%% file: 0_0_abstract.tex
\begin{abstract}
Since the application of Deep Q-Learning to the continuous action domain in Atari-like games, Deep Reinforcement Learning (Deep-RL) techniques for motion control have been qualitatively enhanced. Nowadays, modern Deep-RL can be successfully applied to solve a wide range of complex decision-making tasks for many types of vehicles. Based on this context, in this paper, we propose the use of Deep-RL to perform autonomous mapless navigation for Hybrid Unmanned Aerial Underwater Vehicles (HUAUVs), robots that can operate in both, air or water media. We developed two approaches, one deterministic and the other stochastic. Our system uses the relative localization of the vehicle and simple sparse range data to train the network. We compared our approaches with a traditional geometric tracking controller for mapless navigation. Based on experimental results, we can conclude that Deep-RL-based approaches can be successfully used to perform mapless navigation and obstacle avoidance for HUAUVs. Our vehicle accomplished the navigation in two scenarios, being capable to achieve the desired target through both environments, and even outperforming the geometric-based tracking controller on the obstacle-avoidance capability.
\end{abstract}


%% file: 1_introduction.tex
\section{Introduction}


In the last few years, Deep Reinforcement Learning (Deep-RL) has been increasingly employed in a wide range of fields. Promising results have been achieved in tasks involving control and discrete systems \cite{mnih2013playing}, \cite{schaul2015prioritized}, and continuous systems \cite{lillicrap2015continuous}. In Robotics, Deep-RL was first applied to handling tasks in stable and observable environments \cite{gu2016continuous}. For mobile robots, however, there is a significant increase in complexity due to interactions with barriers in the physical workspace. In this context, Deep-RL approaches typically concentrate on discretizing the problem to solve it \cite{zhu2017target}. 

Interesting results have been achieved in recent works exploring continuous control actions for ground mobile robots \cite{tai2017virtual}, aerial robots \cite{rodriguez2018deep} and underwater robots \cite{carlucho2018}. However, the results are quite limited when considering Hybrid Unmanned Aerial Underwater Vehicles (HUAUVs), since its state-of-art is yet focused on conceptual design, construction, and experimentation of vehicles \cite{drews2014hybrid, neto2015attitude, da2018comparative, maia2017design, lu2019multimodal, ma2018research, mercado2019aerial}. 
Potential applications of HUAUVs range from mapping, industrial inspection, search and rescue, to military-grade applications.



\begin{figure}[tbp!]
\centering
\includegraphics[width=\linewidth]{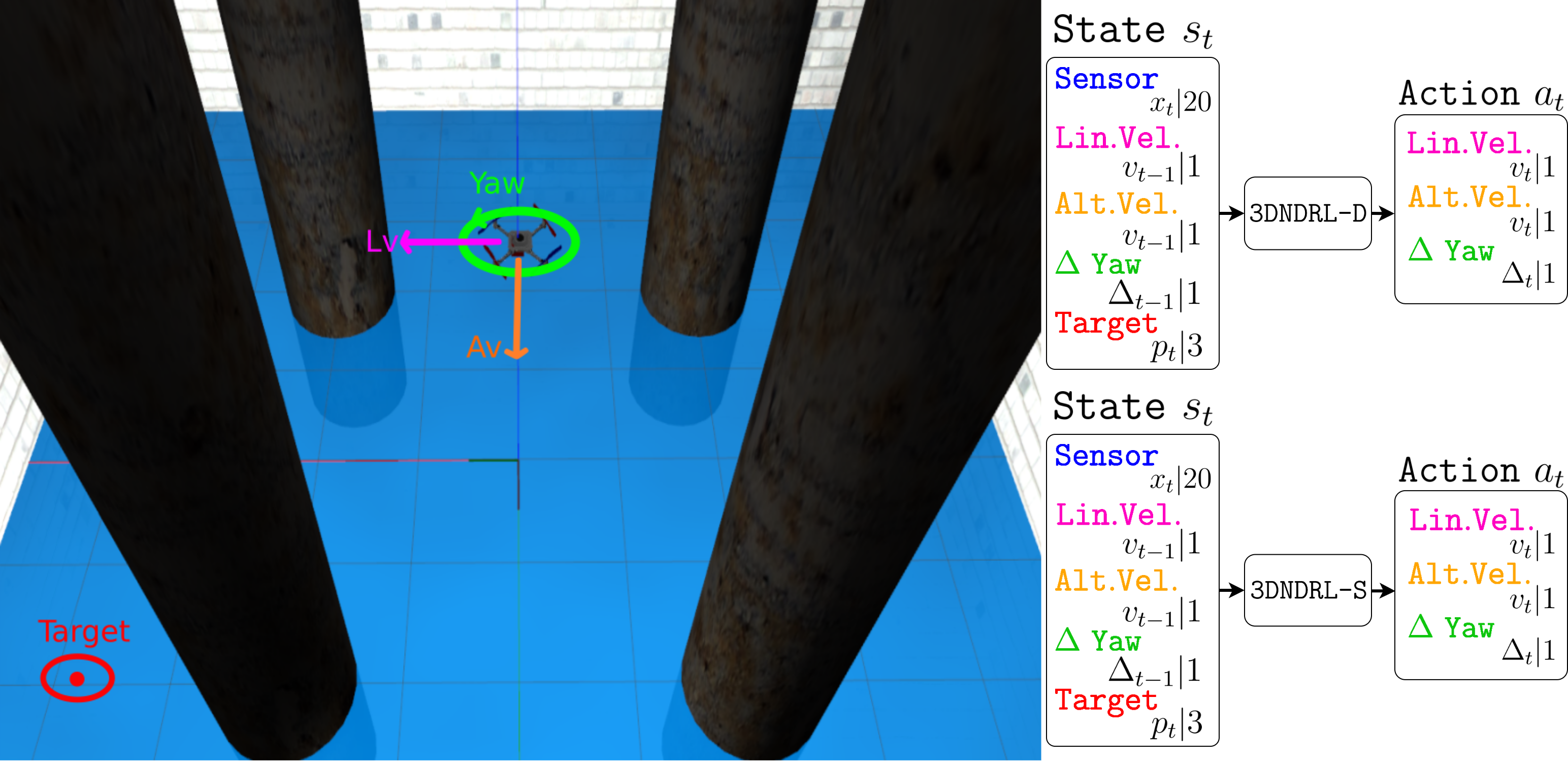}
\caption{Our HUAUV performing among obstacles in one of the scenarios (left) and the model structure with inputs and outputs of our methods, 3DNDRL-D and 3DNDRL-S (right).}
\label{fig:system_def}
\vspace{-5.1mm}
\end{figure}

In this work, we focus on demonstrating the effectiveness of Deep-RL on the mapless navigation of HUAUVs. More specifically, we tackled the task of goal-oriented navigation with medium transition in a two-fold way for the HUAUV presented in \cite{drews2014hybrid}. We proposed two approaches: 
1) A deterministic based on Deep Deterministic Policy Gradient (DDPG);
2) A stochastic based on Soft Actor-Critic (SAC).
We train and test in two different scenarios with both approaches, then compare the results with Lee et al. \cite{lee2010geometric}. Our system definition can be seen in Fig. \ref{fig:system_def}.

This work contains the following contributions:
%


\begin{itemize}
\item We propose two new approaches based on state-of-art Deep-RL algorithms for ground robots that can successfully perform goal-oriented mapless navigation for HUAUVs, using only range data readings and the vehicles' relative localization data.


\item We demonstrate that, with our approaches, the robot is capable to arrive at the desired target avoiding collisions. However, with a geometrically dependent tracking controller, the robot is unable to bypass the drilling risers. We also provide a completely built ROS package with a real-world described HUAUV.

\item Through 3D realistic simulation, we show that our approaches can resolve physical difficulties imposed by an environment submerged in water in the same manner as in an environment surrounded by air with wind and achieve the intended target.

\item We also show that our approaches managed to perform the medium transition between the environments, which is a problem avoided by related works because of its complexity.
\end{itemize}




The works are structured as follows: we discuss the related works in Sec \ref{related_works}. Then, in Sec \ref{methdology}, we present our approaches. The results obtained are presented in Sec \ref{results}. Finally, our contributions and future works are summarized in Sec \ref{conclusion}.

%% file: 2_related_works.tex
\section{Related Works}
\label{related_works}

The current literature of HUAUVs is still focused on design \cite{drews2014hybrid, neto2015attitude, iruthayanathan2016performance, da2018comparative, maia2017design, lu2019multimodal, ma2018research, mercado2019aerial}. However, it does not cover the autonomous navigation problem. The HUAUV described in this work is based on the one proposed by Drews et al. at \cite{drews2014hybrid} and partially extended by Neto et al. \cite{neto2015attitude}.

On the other hand, for the mapless navigation problem, some Deep-RL surveys have already been carried out in robotics tasks, showing how efficiently we can solve the problem by using learning approaches \cite{kober2013reinforcement}, \cite{kormushev2013reinforcement}, \cite{tobin2017domain}.
For example, Tai et al. \cite{tai2017virtual} used 10-dimensional range findings and the relative distance of the vehicle to a target as inputs and continuous steering signals as outputs for a mapless motion planner of a ground robot. The study concluded that a mapless motion planner based on the DDPG algorithm can be successfully trained and used to complete the task of navigating to a target. 
Zhu et al. \cite{zhu2017target}, Chen et al. \cite{chen2017socially}, Ota et al. \cite{ota2020efficient}, de Jesus et al. \cite{jesus2019deep} and others have already successfully applied Deep-RL approaches to perform mapless navigation-related tasks for terrestrial mobile robots.

The use of Deep-RL is more limited when applied to problems related to mapless navigation for Unmanned Aerial Vehicles (UAVs). A DDPG-based approach was used by Rodriguez et al. \cite{rodriguez2018deep} to tackle the landing on a moving platform problem. It used Deep-RL with the RotorS framework \cite{furrer2016rotors} for simulation of UAVs in the Gazebo simulator. Sampedro et al. \cite{sampedro2019fully} suggested a DDPG-based approach for the Search and Rescue task in indoor environments using visual information of real and a simulated UAV. Kang et al. \cite{kang2019generalization} used visual information as well, but concentrated on the problem of collision avoidance. Barros et al. \cite{2020arXiv201002293M} used a SAC-based approach to low-level control of a UAV in a go-to-target task.

Underwater mapless navigation is also performed by a hybrid vehicle, and it has been also addressed with Deep-RL-based approaches. Carlucho et al. \cite{carlucho2018} used a DDPG-based approach to perform the navigation of an Unmanned Underwater Vehicle (UUV). It did so by controlling the vehicles' thrusters and using the submarine-like UUV range findings and the vehicles' relative position as inputs. Yu et al. \cite{yu2017} focused on comparing a DDPG-based approach with a PID control method on a path following task for his described submarine-like UUV in Gazebo. Wu et al. \cite{wu2019} proposed an approach that uses Deep-RL for 2D obstacle avoidance and 3D pipeline follow-up tasks. Zhang et al. \cite{zhang2020} addressed the mapless navigation problem for a submarine-like UUV as well but focused on a Deep-RL interactive approach. All works used the underwater framework for Gazebo, UUV simulator \cite{manhaes2016uuv}.

Our proposal differs from the discussed works by only using the vehicle's relative localization data and not its explicit localization data. We also propose two Deep-RL approaches which we called 3DNDRL-D and 3DNDRL-S, a deterministic one and also a bias-stochastic one to further enhance our work. We compare them with the mapless navigation performed by a traditional geometric tracking controller that can be used for mapless navigation \cite{lee2010geometric}.

%% file: 3_methodology.tex
\section{Methodology}
\label{methdology}

In this section, we discuss our Deep-RL approaches and our vehicle. We detail the network structure for both deterministic and stochastic agents. We also present the proposed reward function for the task the vehicle must accomplish autonomously.

\subsection{Deep Reinforcement Learning}

Deep-RL aims to endow an agent with some capabilities by maximizing a reward function. By setting the agent's actions, the traditional deep Q-network (DQN) algorithm \cite{mnih2013playing} managed to outperform humans in many games. However, it only worked in discrete action spaces, being unable to handle complex observation spaces. Since many tasks in robot control have continuous action spaces and DQN was not designed for continuous domains, another algorithm must be used to deal with such a problem \cite{lillicrap2015continuous}.

\subsubsection{3D Navigation Deep Reinforcement Learning Deterministic}

The DDPG architecture has been proposed as an extension to the DQN, and it became a founding block for Deep-RL applications of mobile robots in continuous observation spaces \cite{jesus2019deep}. It consists of an actor-critic approach that uses approximation functions to learn continuous space policies.
With that in mind, we developed a method based on the DDPG algorithm to develop our deterministic approach called 3DNDRL-D. In this approach, an exploration policy needs to be implemented.
To do this, a sampled noise $\mu'$ is added to the actor-network policy, given the noise process $\mathcal{N}$, such that:

\begin{equation*}
    \mu' = \mu(s_t) + \mathcal{N},
\end{equation*}
%
\noindent where the Ornstein-Uhlenbeck method is typically chosen for physical control problems \cite{uhlenbeck1930theory}. 

A replay memory is suggested during the training \cite{schaul2015prioritized}. It means saving the results of the exploration state $s_t$ during the episode, as well as the action $a_t$, the reward $r_t$, and the next state $s_{t+1}$, to sample them randomly in the future. 
A $50000$-step buffer replay memory was used in this work in both Deep-RL approaches. 


\subsubsection{3D Navigation Deep Reinforcement Learning Stochastic}

For the 3DNDRL-S approach, we proposed a new method based on the SAC algorithm \cite{haarnoja2018soft}. SAC consists of a bias-stochastic actor-critic that, using approximation functions, learns continuous action space policies. Providing a bridge between optimizing stochastic policy and DDPG-style approaches, it has been built based on two neural networks: an actor-network and a critic-network. These networks predict the behavior of the current state, alongside a temporal-difference error signal at each point of time.

Besides focusing only on maximizing the reward of the scheme, this algorithm attempts to increase the policy's entropy, which tells how unpredictable a variable is. The highest the entropy, the highest the motivation is for the robot to explore new states.
A variable that has never changed its value has zero entropy and no incentive at all to be modified. This motivation is achieved by assigning equal probability to the behavior of identical Q-value and by ensuring that it does not fail by a single uncertain action in the approximated Q-function.

\subsection{Simulated Environments}


To broaden the use of the Gazebo simulator for several types of robots in distinct environments, many applications have been developed. Among them, the RotorS is a framework that has made it possible to examine aerial vehicles' behavior. Specifically, a geometric tracking controller \cite{lee2010geometric} was implemented in RotorS, which provides access to different command levels, such as angular rates, attitude, location control, and even goal-oriented mapless navigation. It also enables the simulation of wind, which was done with an Ornstein-Uhlenbeck noise method.

Another framework used is the UUV simulator \cite{manhaes2016uuv}. This Gazebo extension enables the simulation of hydrodynamic and hydrostatic effects, as well as sensors, thrusters, and external perturbations. It is possible to set the vehicle's underwater model, with parameters such as the center of buoyancy, additional mass, volume, etc, as well as the characteristics of the underwater environment itself, such as water flow velocity, for example. Unlike RotorS, UUV sim does have a smaller number of described vehicles. 

To the best of our knowledge, our described HUAUV is the first based on a quadrotor structure already developed for the Gazebo simulator. 
We simulated our HUAUV with these frameworks and other available tools, built environments to train and evaluate our approach and compare our agents' results with the geometric tracking controller.

We developed two environments that simulate a water tank. The first environment is a 10$\times$10$\times$6 meters water tank with a free area for the robot to move, where the walls are the only things that the robot can collide. The water column is one-meter height inside the tank. In case of collision, a negative reward is given for the current episode, stopping the episode. The second environment has the water tank plus 4 fixed obstacles that simulate drilling risers. In this more complex environment, the vehicle may have to outline the drilling risers to reach a goal. The second environment can be seen in Figure \ref{fig:system_def}.


\subsection{Vehicle Description}

Using the RotorS and UUV simulator frameworks, as well as the Deep-RL structure proposed in this work, we created a ROS package\footnote{\scalebox{0.92}{\url{https://github.com/ricardoGrando/hydrone\_deep\_rl\_icra}}} containing the description of our vehicle \cite{drews2014hybrid} with some of its latest modeled information \cite{horn2019study}. We defined it using the actual dimensions and constants of our real vehicle, such as inertia, motor coefficients, mass, rotor velocity, and others. We also used the mesh given by a modeling software that was used to design it as visual detail. 
%

\begin{figure}[ht]
\vspace{-5mm}
  \subfloat[Real vehicle.\label{fig:real_v}]{
	\begin{minipage}[c][0.60\width]{
	   0.23\textwidth}
	   \centering
	   \includegraphics[width=0.79\textwidth]{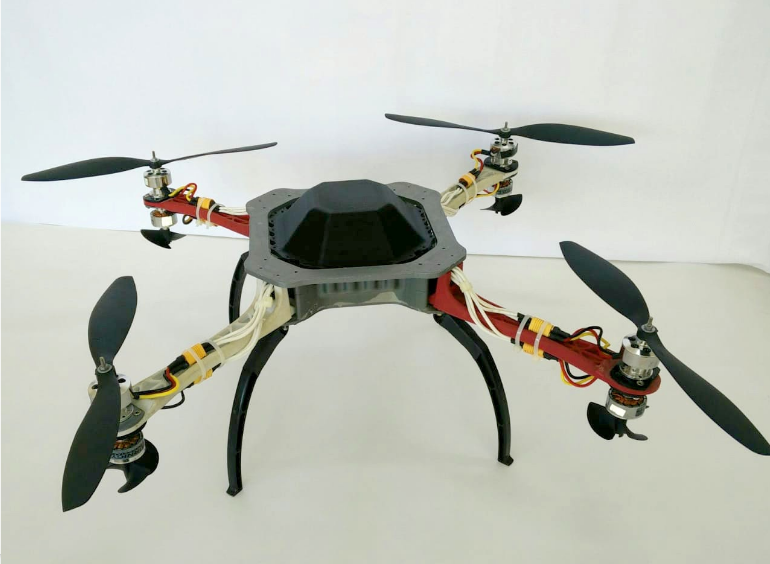}
	\end{minipage}}
 \hfill 	
  \subfloat[Described vehicle.\label{fig:desc_v}]{
	\begin{minipage}[c][0.60\width]{
	   0.23\textwidth}
	   \centering
	   \includegraphics[width=\textwidth]{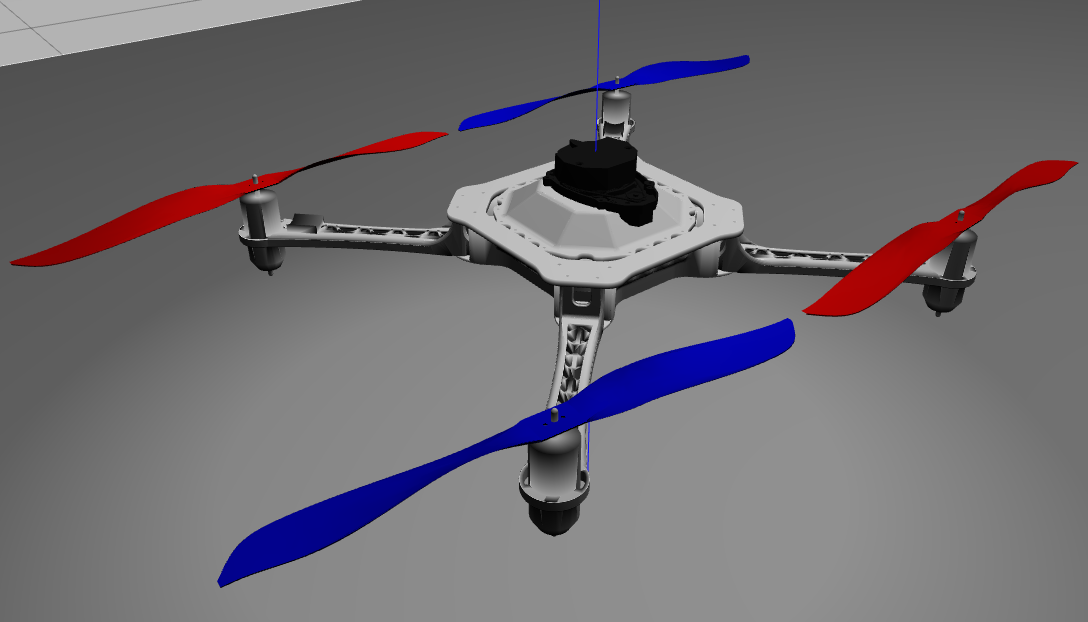}
	\end{minipage}}
\caption{Simulated vehicle based on a real-world HUAUV.}
\label{fig:vehicles}
\vspace{-2mm}
\end{figure}

The hydrodynamic of the vehicle was also estimated using modeling software. Fig. \ref{fig:vehicles} shows the real vehicle and the described vehicle with a simulated range findings sensor attached. In the real vehicle, the range findings readings for both air and underwater environments can be obtained using sensors like a lidar and a Sonar, respectively. As well as for the vehicle's localization data, which could be obtained by a combination of sensors like GPS and USBL for example.

We used the geometric tracking controller implementation of the RotorS to set the $\Delta$ yaw and linear velocity in the vehicle. We built an internal plugin that receives and sets the linear velocity and the $\Delta$ yaw in the controller. In the conversion from polar to Cartesian coordinates, the linear velocity constant obtained from the network output is set to the vehicle $x$ and $y$ linear velocities according to its yaw angle plus $\Delta$ yaw angle from the action. The mapless navigation performed by the robots was compared with the RotorS' geometric tracking controller for goal-oriented mapless navigation.

\subsection{Proposed Approaches}

In this work, we aim 
to navigate our described HUAUV autonomously from a starting point to a target point in a different environment without any environmental knowledge, preventing any collision with the scenario by using only range readings and the vehicle localization data. Its translation function, therefore, is defined as:

\begin{equation*}
    s_t = f(x_t, t_t, m_{t-1}),
\end{equation*}
\noindent where $x_t$ is the range findings, $t_t$ is the target's relative information, and $m_{t-1}$ is the vehicle's motion of the last step. These variables describe the current state of $s_t$ of the vehicle. Used in many Deep-RL works for terrestrial mobile robots \cite{tai2017virtual}, \cite{jesus2019deep}, this model helps to obtain the behavior of the agent given its current state.

\subsubsection{Networks Structure}

\begin{figure}[ht]
    \vspace{2.5mm}
    \centering
    \includegraphics[scale=0.13]{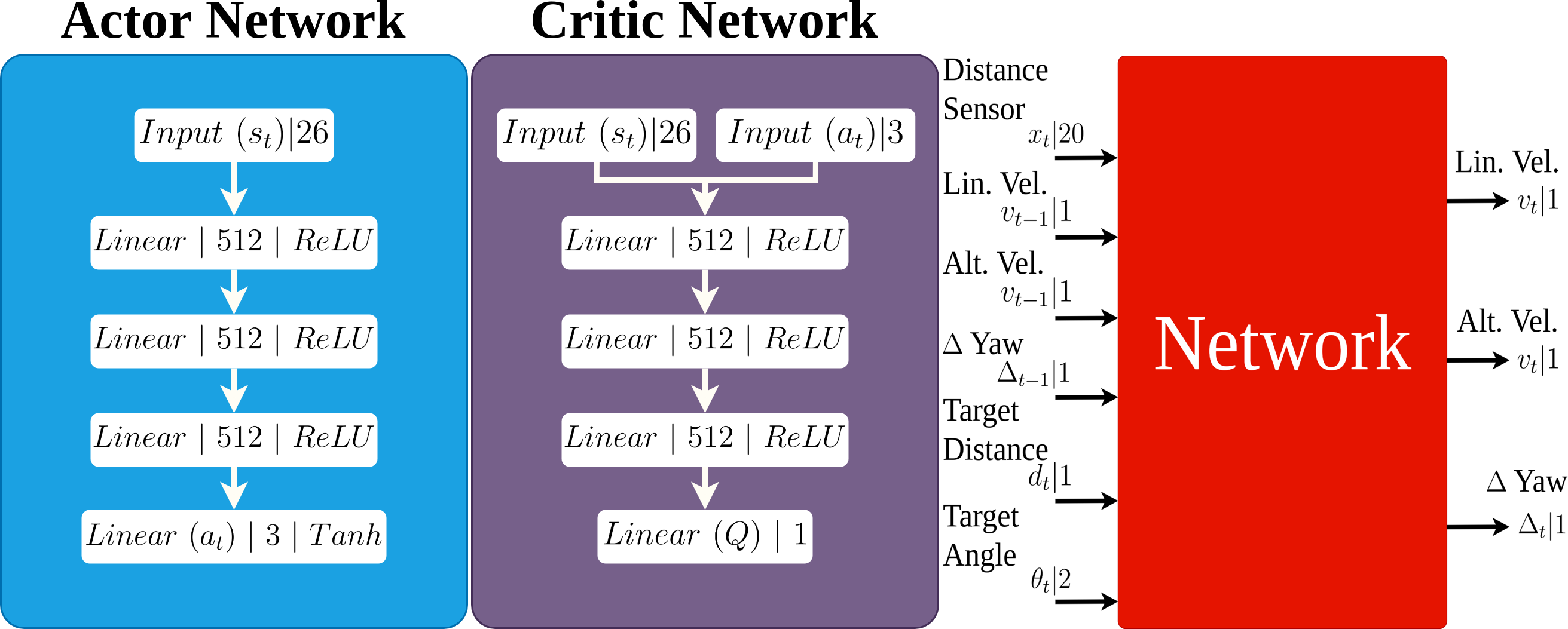}
    \caption{Deep-RL models structures (left) and network inputs and outputs (right).}
    \label{fig:network_Strcuture}
    \vspace{-6mm}
\end{figure}

As shown in Fig. \ref{fig:network_Strcuture}, the network has 26 inputs and 3 outputs, for both 3DNDRL-D and 3DNDRL-S approaches. Inputs are 20 range findings, the previous linear and altitude velocities and $\Delta$ yaw, the vehicle's relative position to the target, and relative angles to the target in the x-y plan and in the z-distance plan. The samples from the sensor range between $-135\degree$ and $135\degree$, equally spaced by $13.5\degree$. The input angles are used to force the vehicle to minimize it and improve the knowledge. As well as the target distance that encourages the network to learn how to minimize it. 
On the other hand, outputs are the linear velocity and the $\Delta$ yaw to control the vehicle.

The networks of our proposed 3DNDRL-D and 3DNDRL-S approaches have 3 hidden, fully-connected layers with 512 neurons each and connected through ReLU activation for the actor-network, which uses the current state as input. The action ranges between $-1$ and $1$ and the hyperbolic tangent function (\textit{Tanh}) was used as the activation function. The outputs are scaled between $0$ and $0.25$ $m/s$ for the linear velocity, from $-0.25$ $m/s$ to $0.25$ $m/s$ for the altitude velocity and from $-0.25$ to $0.25$ $rad$ for the $\Delta$ yaw. The 3DNDRL-D and 3DNDRL-S network structure model used can be seen in Fig. \ref{fig:network_Strcuture}. For both approaches, the Q-value of the current state is predicted in the critic network, while the actor-network predicts the current state. 



\subsubsection{Reward Function}

A reward and penalty function must be established for both Deep-RL strategies. Rewards and penalties are values assigned to the performance of the agent, allowing the agent to learn hyper-parameters through the feed-forward and back-propagation phases of the networks. It was used a very simple rewarding and penalty strategy described as follows:

\begin{equation}
r(s_t, a_t)= 
\begin{cases}
    r_{arrive}           & \text{if } d_t < c_d\\
    r_{collide}          & \text{if } min_x < c_o\\
\end{cases}
\end{equation}

If the robot hits the goal within a $c_d$ meter margin, which was set as $0.25$ meters, a positive $r_{arrive}$ reward of $100$ is given. If the vehicle collides with the wall of the water tank or one of the drilling risers a negative reward of $-10$ is granted to the episode. The collision is verified by testing if the minimum distance of the 20 range readings is less than a $c_o$ distance of $0.5$ meters.

%% file: 4_experimental_results.tex
\section{Experimental Results}
\label{results}

\subsection{Experimental setup}

The whole system was implemented using ROS and Gazebo frameworks. The Deep-RL approaches were implemented using Python programming language, while the vehicles' related plugins were partially implemented in C++ and Python. The implementation of the neural networks was carried out with the PyTorch\footnote{\url{https://pytorch.org/}} library. 
The performance of our approaches can also be observed in a complementary video\footnote{\url{https://youtu.be/EGY3xldWCdA}}.

\subsection{Training setup}

An agent was trained for each one of the two scenarios and each one of the proposed approaches. For all the training episodes, the vehicle's initial position was set as (0.0, 0.0, 2.5), with its front part pointing towards the positive $x$-axis. A randomly generated target which the vehicle must reach was created at each episode. The episode ends if the vehicle collides with the water tank's wall or with one of the drilling risers or when a total amount of 500 steps has been reached in that episode. It means that the episode does not end if the vehicle reaches the goal. Instead, a new random target is generated during the same episode. As consequence, the total amount of reward of a given episode can surpass the $100$ reward set for navigating properly. For the second scenario, specifically, no target goals inside the obstacles boundaries were generated to avoid creating a non-reachable target.  

The actor and critic neural networks have been trained with Adam optimizer and with a learning rate of $10^{-3}$ for both approaches. The selected minibatch size was 256 for both approaches as well. We set a limit of training of $1000$ episodes for the first scenario and $2500$ episodes for the second scenario. The limits were defined empirically since good results could be observed around these episodes. 

\subsection{Results}

\begin{figure}[h]
\vspace{-5mm}
  \subfloat[First environment.\label{fig:reward_1}]{
	\begin{minipage}[c][0.72\width]{
	   0.235\textwidth}
	   \centering
	   \includegraphics[width=1.0\textwidth]{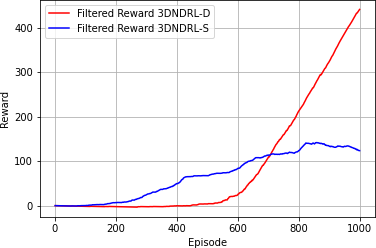}
	\end{minipage}}
 \hfill 	
  \subfloat[Second environment.\label{fig:reward_2}]{
	\begin{minipage}[c][0.72\width]{
	   0.235\textwidth}
	   \centering
	   \includegraphics[width=1.0\textwidth]{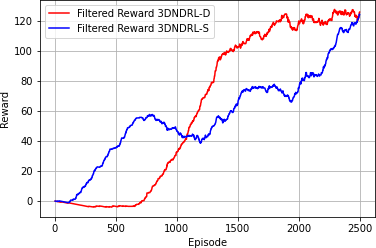}
	\end{minipage}}
\caption{Moving average of the reward over 300 episodes of the training.}
\label{fig:rewards}
\vspace{-3mm}
\end{figure}

The moving average of the reward by each approach in each scenario can be seen in Fig. \ref{fig:rewards}. One can observe that in the first scenario the 3DNDRL-D approach took more episodes than the 3DNDRL-S to start learning the task, but yielded a considerably higher amount of reward at the end of the training. In the second scenario, however, both approaches yielded a similar average of rewards at the finish. So, it is expected that the 3DNDRL-D presents better results in the first scenario, while a similar performance was obtained in the second and more complex scenario.

\begin{figure*}[b]
\vspace{-6mm}
  \centering
  \subfloat[3DNDRL-D, first env, air-water.\label{fig:cmd_ddpg_1_air_to_water}]{
	\begin{minipage}[c][0.65\width]{0.242\textwidth}
	   \centering
	   \includegraphics[width=\textwidth]{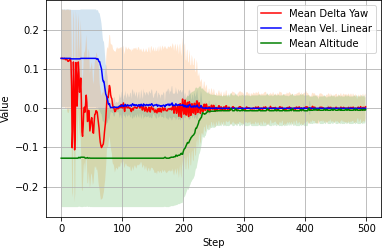}
	\end{minipage}}
 \hfill 	
  \subfloat[3DNDRL-S, first env, air-water.\label{fig:cmd_sac_1_air_to_water}]{
	\begin{minipage}[c][0.65\width]{0.242\textwidth}
	   \centering
	   \includegraphics[width=\textwidth]{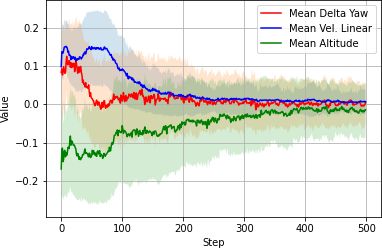}
	\end{minipage}}
  \hfill 	
  \subfloat[3DNDRL-D, second env, air-water.\label{fig:cmd_ddpg_2_air_to_water}]{
	\begin{minipage}[c][0.65\width]{0.242\textwidth}
	   \centering
	   \includegraphics[width=\textwidth]{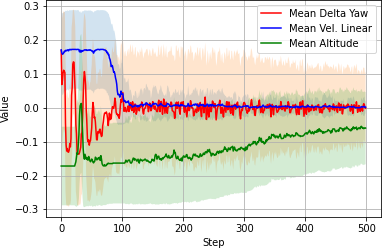}
	\end{minipage}}
  \hfill 	
  \subfloat[3DNDRL-S, second env, air-water.\label{fig:cmd_sac_2_air_to_water}]{
	\begin{minipage}[c][0.65\width]{0.242\textwidth}
	   \centering
	   \includegraphics[width=\textwidth]{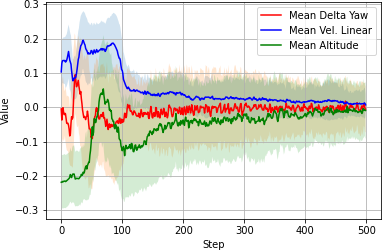}
	\end{minipage}}
  \hfill 
  \vspace{-1mm}
  \subfloat[3DNDRL-D, first env, water-air.\label{fig:cmd_ddpg_1_water_to_air}]{
	\begin{minipage}[c][0.65\width]{0.242\textwidth}
	   \centering
	   \includegraphics[width=\textwidth]{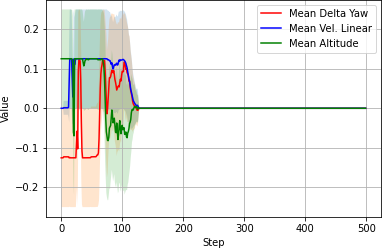}
	\end{minipage}}
 \hfill 	
  \subfloat[3DNDRL-S, first env, water-air.\label{fig:cmd_sac_1_water_to_air}]{
	\begin{minipage}[c][0.65\width]{0.242\textwidth}
	   \centering
	   \includegraphics[width=\textwidth]{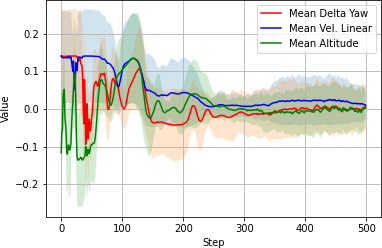}
	\end{minipage}}
  \hfill 	
  \subfloat[3DNDRL-D, second env, water-air.\label{fig:cmd_ddpg_2_water_to_air}]{
	\begin{minipage}[c][0.65\width]{0.242\textwidth}
	   \centering
	   \includegraphics[width=\textwidth]{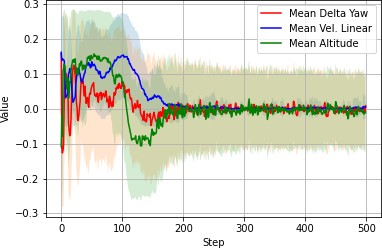}
	\end{minipage}}
  \hfill 	
  \subfloat[3DNDRL-S, second env, water-air.\label{fig:cmd_sac_2_water_to_air}]{
	\begin{minipage}[c][0.65\width]{0.242\textwidth}
	   \centering
	   \includegraphics[width=\textwidth]{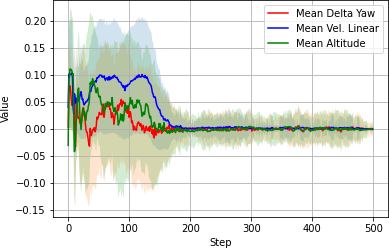}
	\end{minipage}}
  \hfill
  \caption{Mean and standard deviation of agents' behavior in each simulated scenario over 100 navigation trials.}
  \label{fig:cmd}
  \vspace{-2mm}
\end{figure*}

We addressed the task of goal-oriented navigation in a two-fold way for each scenario and each approach. In the first one, the vehicle was set to start in the air and diving to an underwater target, while in the second one it was placed underwater and should navigate upwards to an aerial target. We set a fixed starting position and a target location to test our agents and compare their performance with Lee et al. \cite{lee2010geometric}. In the first scenario, the initial position was (0.0, 0.0, 2.5) in the Gazebo cartesian coordinates, while the target position was (2.0, 3.0, -1.0) at the bottom of the water tank, representing a transition from air to water. 

For the medium transition water to air, the coordinates were inverted, with (0.0, 0.0, 2.5) as the target and (2.0, 3.0, -1.0) as the initial position. For the second scenario, the same was done. The coordinates (0.0, 0.0, 2.5) and (3.6, -2.4, -1.0) were defined. In this scenario, the coordinates were defined in such a way that drilling risers were in the way. So, besides navigating through the distinct environments, the vehicle should outline the obstacle to reach the goal without a collision as well. The navigating performance was measured with a small subset of 100 testing trials for each context, and a comparative metric is shown in Table \ref{table:mean_std}.

\begin{table}[ht]
\vspace{-2mm}
\centering
\setlength{\tabcolsep}{3.5pt}
\caption{Mean and standard deviation metrics over 100 navigation trials in two different simulated scenarios, for both 3DNDRL-D, 3DNDRL-S, and geometric controller methods.}
\label{table:mean_std}
\begin{tabular}{c c c c c} 
\toprule
Env & Test & $t_{air}$ (s) & $t_{under}$ (s) & Success \\
\midrule
1 & Air-Water-3DNDRL-D & $15.29$ $\pm$ $2.40$ & $33.05$ $\pm$ $10.91$ & $96$ \\
1 & Air-Water-3DNDRL-S & $39.81$ $\pm$ $23.16$ & $23.64$ $\pm$ $19.63$ & $72$ \\
1 & Air-Water-LEE & $3.73$ $\pm$ $0.42$ & $39.21$ $\pm$ $2.94$ & $100$ \\
1 & Water-Air-3DNDRL-D & $18.78$ $\pm$ $1.21$ & $6.10$ $\pm$ $0.17$ & $97$ \\
1 & Water-Air-3DNDRL-S & $33.98$ $\pm$ $26.15$ & $17.36$ $\pm$ $12.87$ & $75$ \\
1 & Water-Air-LEE & $25.50$ $\pm$ $0.27$ & $2.65$ $\pm$ $0.15$ & $100$ \\
2 & Air-Water-3DNDRL-D & $20.06$ $\pm$ $13.92$ & $59.83$ $\pm$ $22.49$ & $54$ \\
2 & Air-Water-3DNDRL-S & $60.88$ $\pm$ $30.25$ & $17.38$ $\pm$ $16.10$ & $37$ \\
2 & Air-Water-LEE & $6.77$ $\pm$ $0.32$ & $0.90$ $\pm$ $0.17$ & $0$ \\
2 & Water-Air-3DNDRL-D & $53.56$ $\pm$ $31.28$ & $5.98$ $\pm$ $1.31$ & $57$ \\
2 & Water-Air-3DNDRL-S & $29.98$ $\pm$ $14.48$ & $6.61$ $\pm$ $0.82$ & $71$ \\
2 & Water-Air-LEE & $10.89$ $\pm$ $1.28$ & $3.64$ $\pm$ $0.58$ & $0$ \\
\bottomrule
\end{tabular}
\vspace{-3mm}
\end{table}

\begin{figure*}[ht]
\centering
  \subfloat[3DNDRL-D behavior in transition air-water. \label{fig:ddpg_1_air_to_water}]{
	\begin{minipage}[c][0.73\width]{0.30\textwidth}
	   \centering
	   \includegraphics[width=\textwidth]{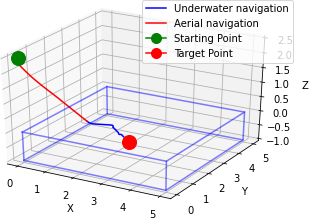}
	\end{minipage}}
 \hfill 	
  \subfloat[3DNDRL-S behavior in transition air-water. \label{fig:sac_1_air_to_water}]{
	\begin{minipage}[c][0.73\width]{0.30\textwidth}
	   \centering
	   \includegraphics[width=\textwidth]{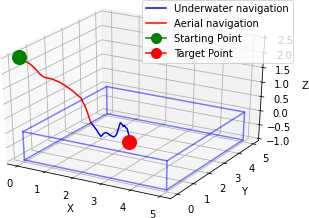}
	\end{minipage}}
 \hfill	
 \subfloat[Lee behavior in transition air-water. \label{fig:lee_1_air_to_water}]{
	\begin{minipage}[c][0.73\width]{0.30\textwidth}
	   \centering
	   \includegraphics[width=\textwidth]{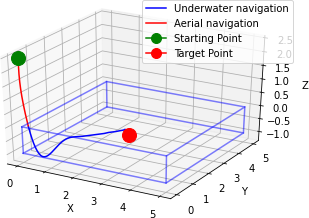}
	\end{minipage}}
 \hfill
 \vspace{-3mm}
 \subfloat[3DNDRL-D behavior in transition water-air. \label{fig:ddpg_1_water_to_air}]{
	\begin{minipage}[c][0.73\width]{0.30\textwidth}
	   \centering
	   \includegraphics[width=\textwidth]{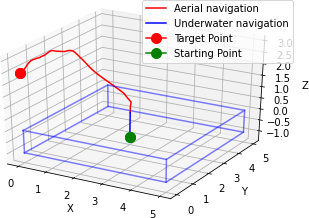}
	\end{minipage}}
 \hfill 	
  \subfloat[3DNDRL-S behavior in transition water-air. \label{fig:sac_1_water_to_air}]{
	\begin{minipage}[c][0.73\width]{0.30\textwidth}
	   \centering
	   \includegraphics[width=\textwidth]{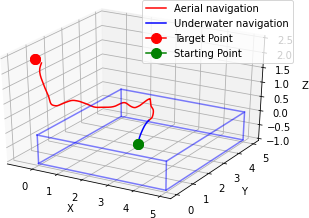}
	\end{minipage}}
 \hfill	
 \subfloat[Lee behavior in transition water-air. \label{fig:lee_1_water_to_air}]{
	\begin{minipage}[c][0.73\width]{0.30\textwidth}
	   \centering
	   \includegraphics[width=\textwidth]{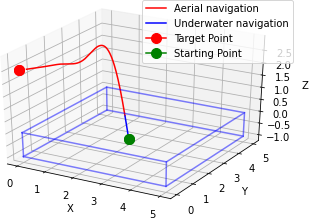}
	\end{minipage}}
 \hfill
 \vspace{-4mm}
    \caption{Behavior sample of each approach tested in the first environment.}
    \label{fig:traj_nav_1}
    \vspace{-6mm}
\end{figure*}

\begin{figure*}[ht]
 \subfloat[3DNDRL-D, air-water. \label{fig:ddpg_2_air_to_water}]{
	\begin{minipage}[c][0.73\width]{0.241\textwidth}
	   \centering
	   \includegraphics[width=\textwidth]{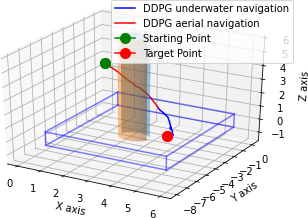}
	\end{minipage}}
 \hfill	
 \subfloat[3DNDRL-D, air-water, x-y plane. \label{fig:ddpg_2_air_to_water_2d}]{
	\begin{minipage}[c][0.73\width]{0.241\textwidth}
	   \centering
	   \includegraphics[width=\textwidth]{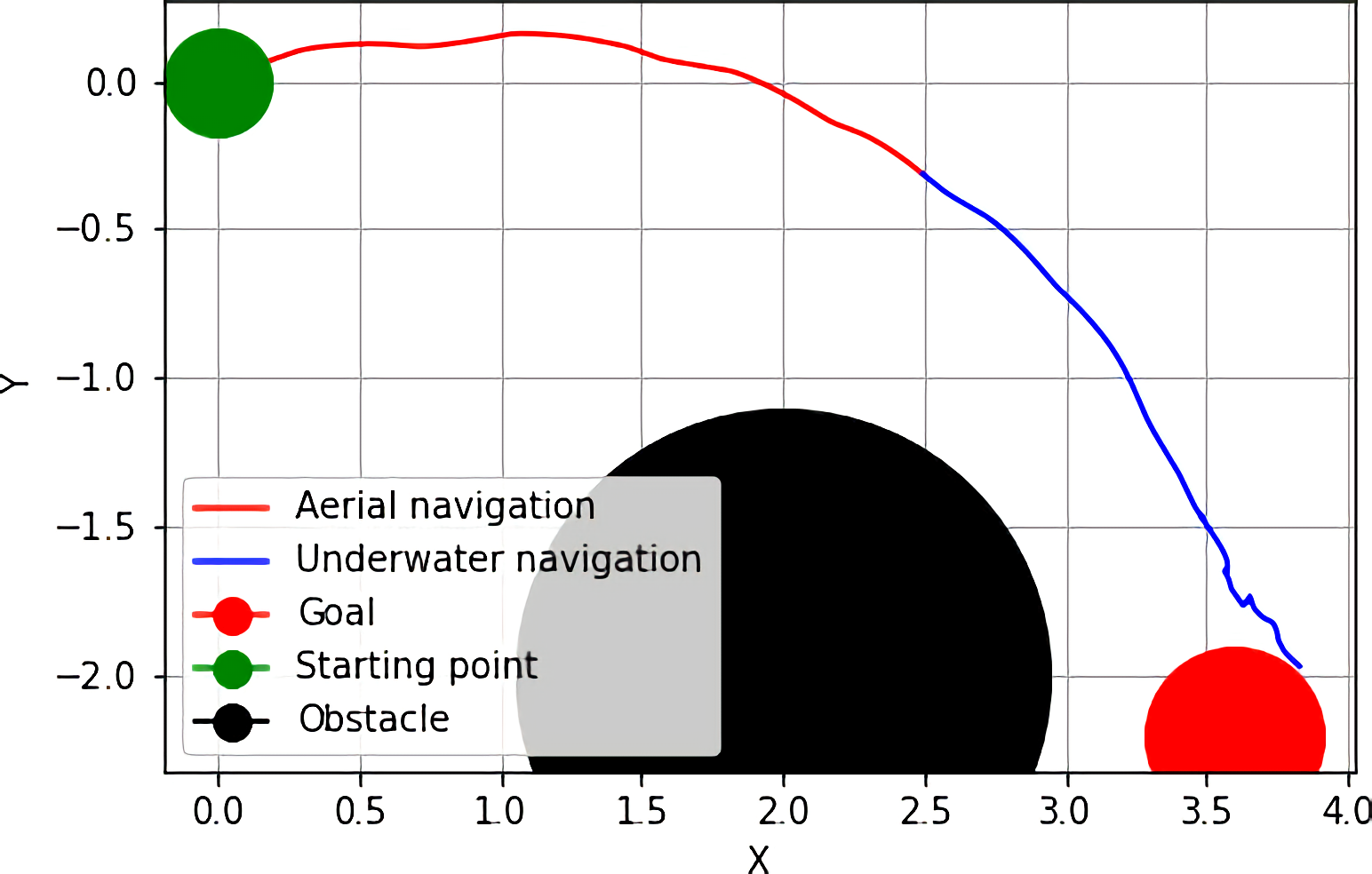}
	\end{minipage}}
 \hfill	
 \subfloat[3DNDRL-D, water-air. \label{fig:ddpg_2_water_to_air}]{
	\begin{minipage}[c][0.73\width]{0.241\textwidth}
	   \centering
	   \includegraphics[width=\textwidth]{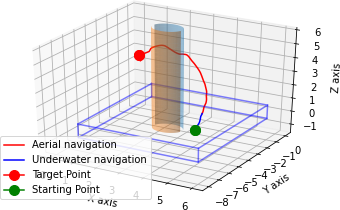}
	\end{minipage}}
 \hfill	
 \subfloat[3DNDRL-D, water-air, x-y plane. \label{fig:ddpg_2_water_to_air_2d}]{
	\begin{minipage}[c][0.73\width]{0.241\textwidth}
	   \centering
	   \includegraphics[width=\textwidth]{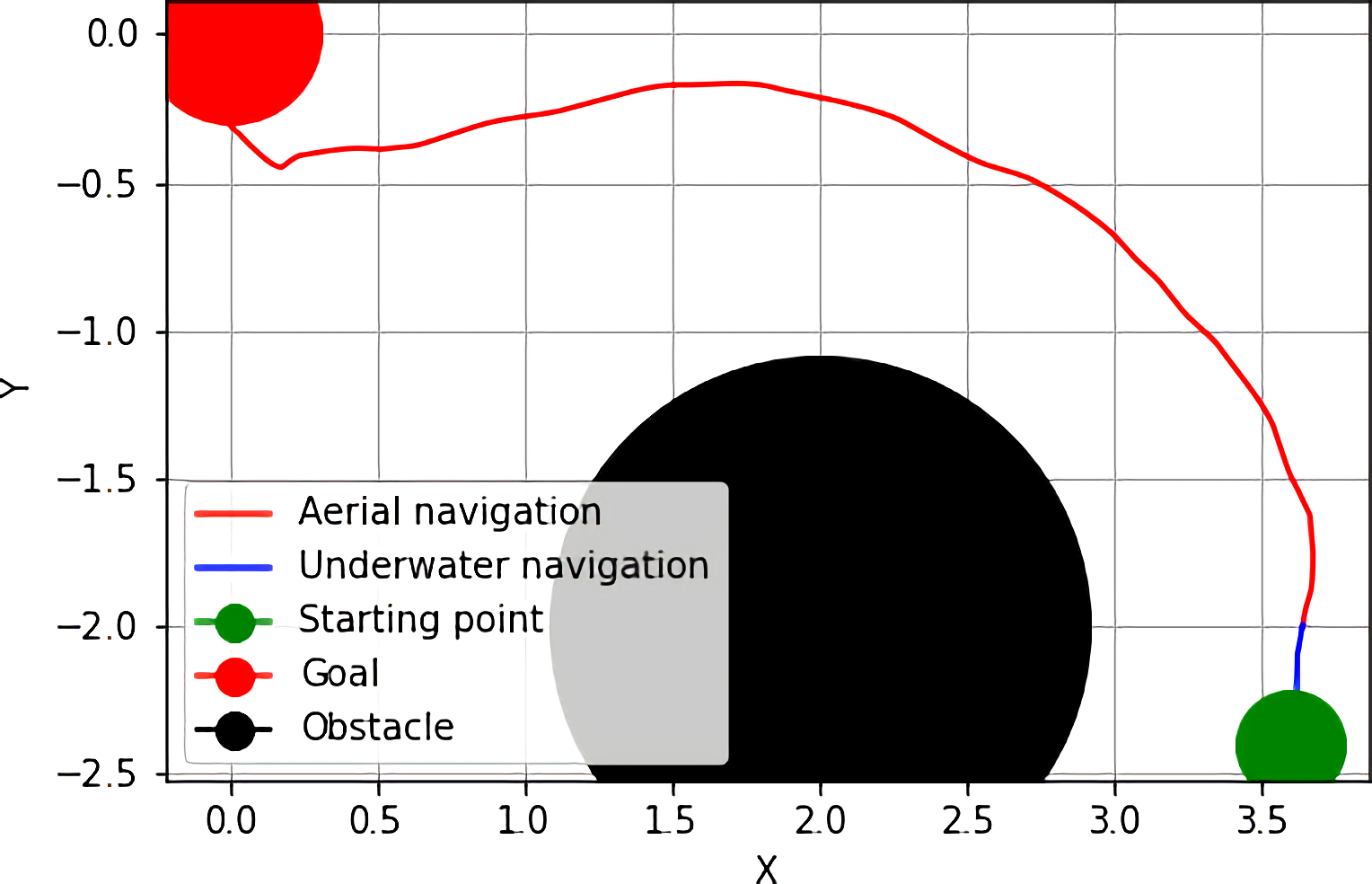}
	\end{minipage}}
 \hfill
 \vspace{-3mm}
 \subfloat[3DNDRL-S, air-water. \label{fig:sac_2_air_to_water}]{
	\begin{minipage}[c][0.73\width]{0.241\textwidth}
	   \centering
	   \includegraphics[width=\textwidth]{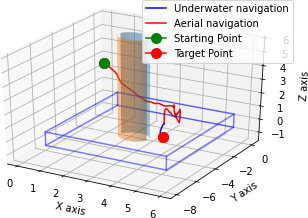}
	\end{minipage}}
 \hfill
 \subfloat[3DNDRL-S, air-water, x-y plane. \label{fig:sac_2_air_to_water_2d}]{
	\begin{minipage}[c][0.73\width]{0.241\textwidth}
	   \centering
	   \includegraphics[width=\textwidth]{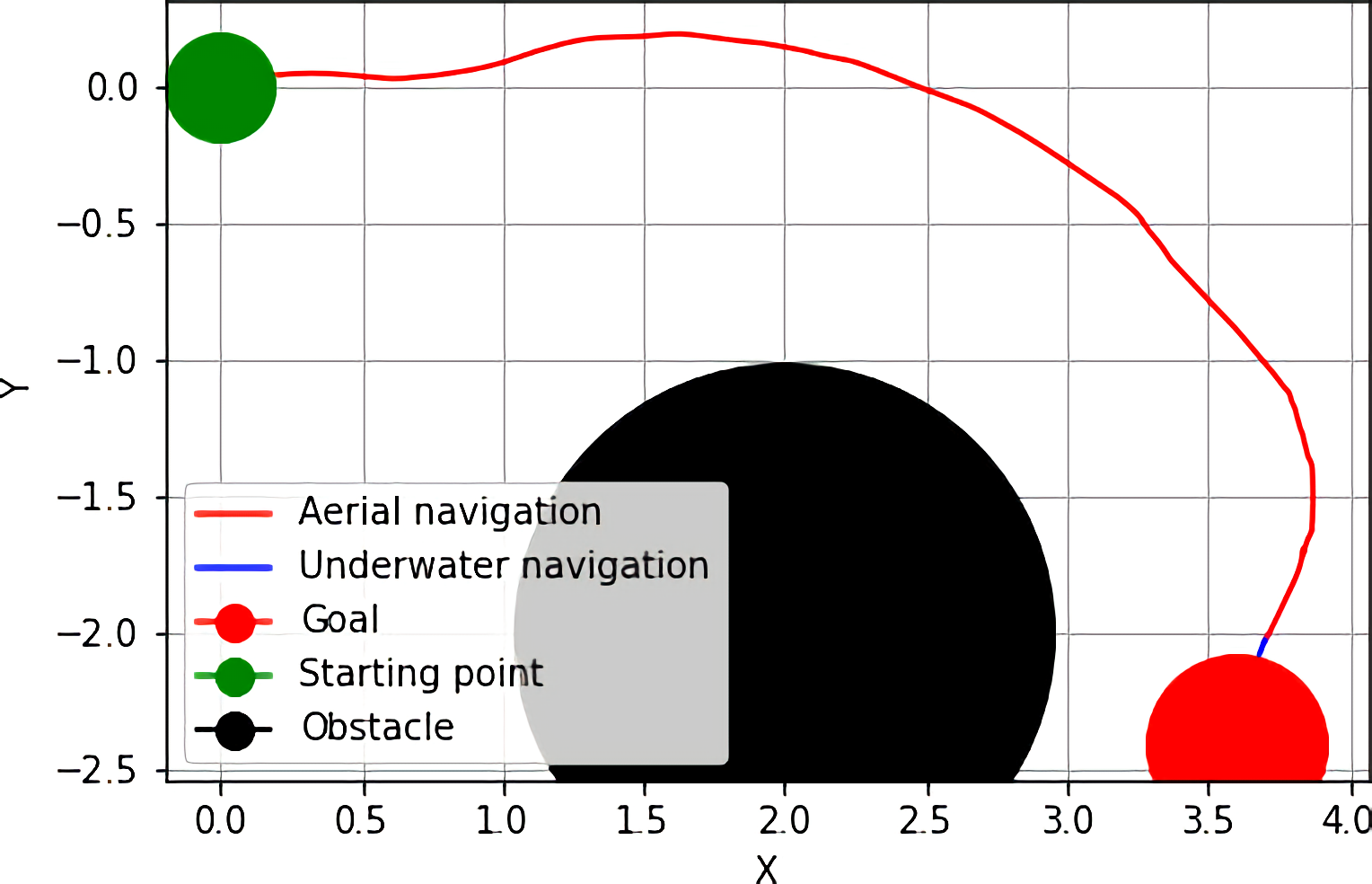}
	\end{minipage}}
 \hfill
 \subfloat[3DNDRL-S, water-air. \label{fig:sac_2_water_to_air}]{
	\begin{minipage}[c][0.73\width]{0.241\textwidth}
	   \centering
	   \includegraphics[width=\textwidth]{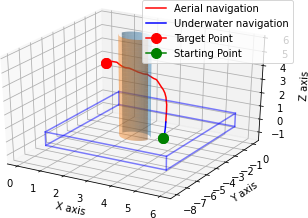}
	\end{minipage}}
 \hfill
 \subfloat[3DNDRL-S, water-air, x-y plane. \label{fig:sac_2_water_to_air_2d}]{
	\begin{minipage}[c][0.73\width]{0.241\textwidth}
	   \centering
	   \includegraphics[width=\textwidth]{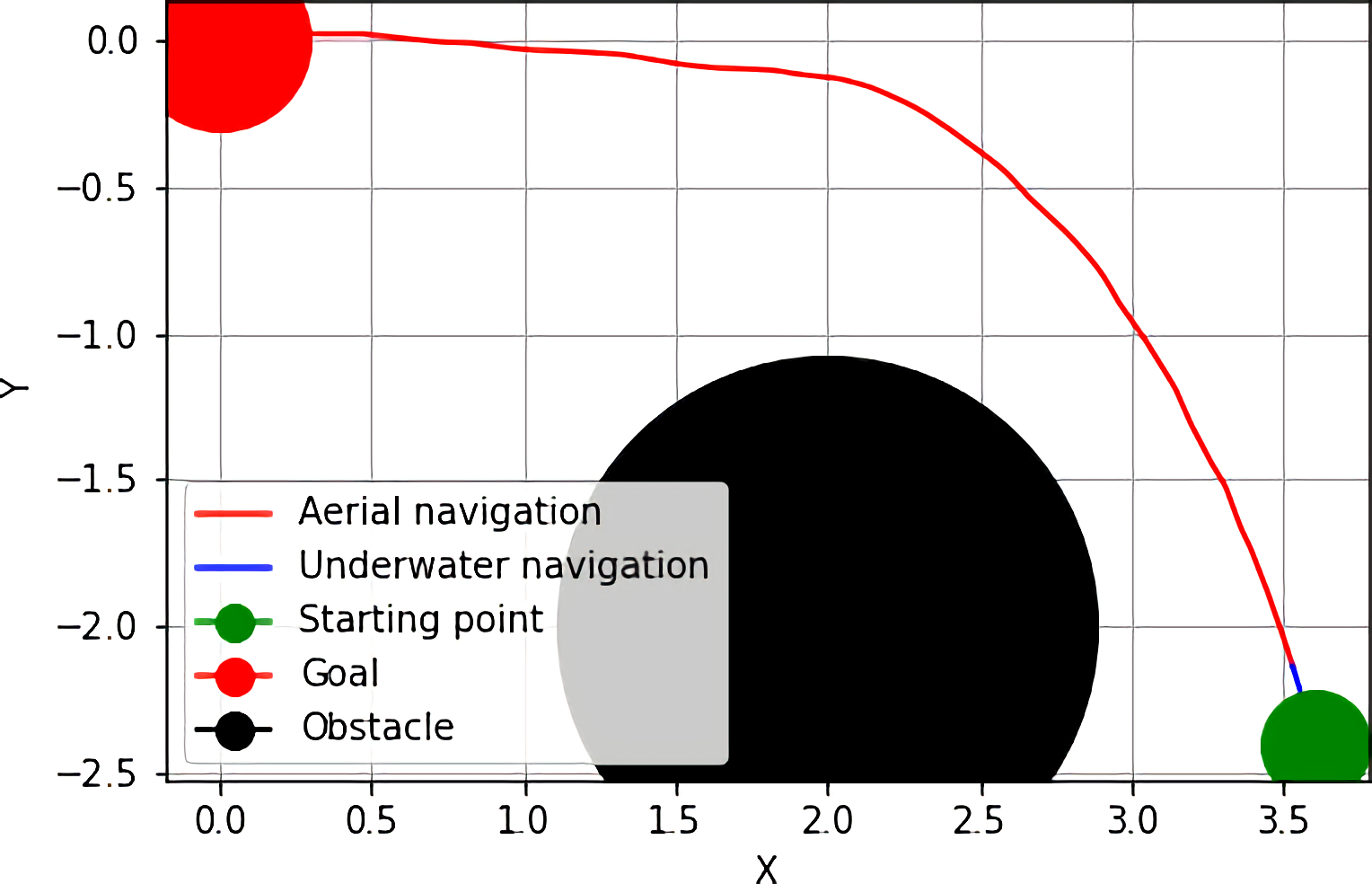}
	\end{minipage}}
 \hfill
 \vspace{-3mm}
 \subfloat[Lee, air-water. \label{fig:lee_2_air_to_water}]{
	\begin{minipage}[c][0.73\width]{0.241\textwidth}
	   \centering
	   \includegraphics[width=\textwidth]{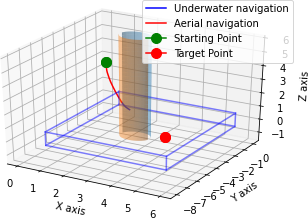}
	\end{minipage}}
 \hfill
 \subfloat[Lee, air-water, x-y plane. \label{fig:lee_2_air_to_water_2d}]{
	\begin{minipage}[c][0.73\width]{0.241\textwidth}
	   \centering
	   \includegraphics[width=\textwidth]{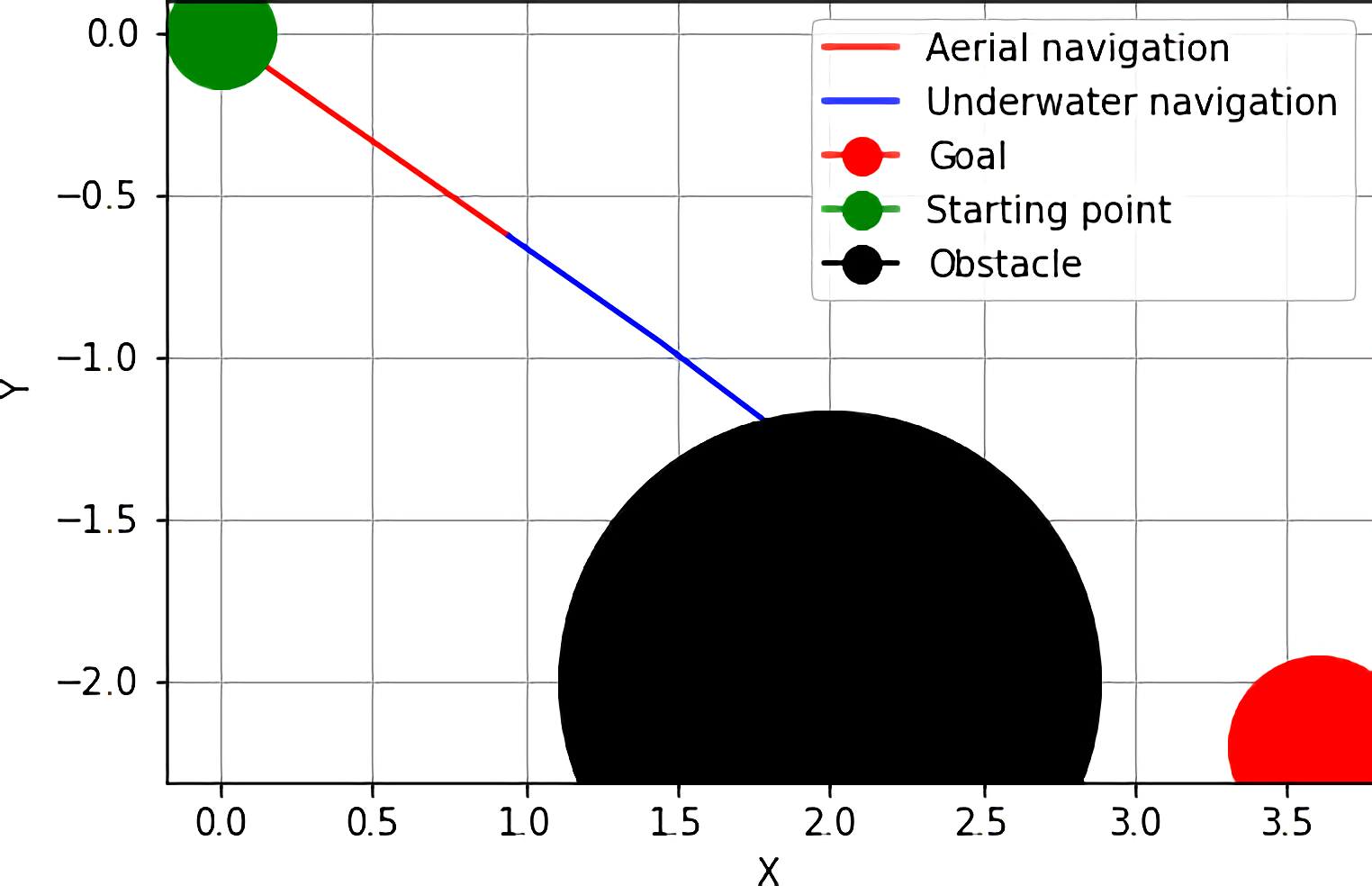}
	\end{minipage}}
 \hfill
 \subfloat[Lee, water-air. \label{fig:lee_2_water_to_air}]{
	\begin{minipage}[c][0.73\width]{0.241\textwidth}
	   \centering
	   \includegraphics[width=\textwidth]{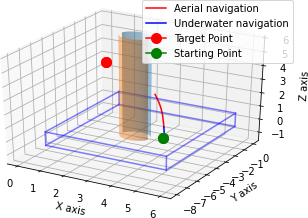}
	\end{minipage}}
 \hfill
 \subfloat[Lee, water-air, x-y plane. \label{fig:lee_2_water_to_air_2d}]{
	\begin{minipage}[c][0.73\width]{0.241\textwidth}
	   \centering
	   \includegraphics[width=\textwidth]{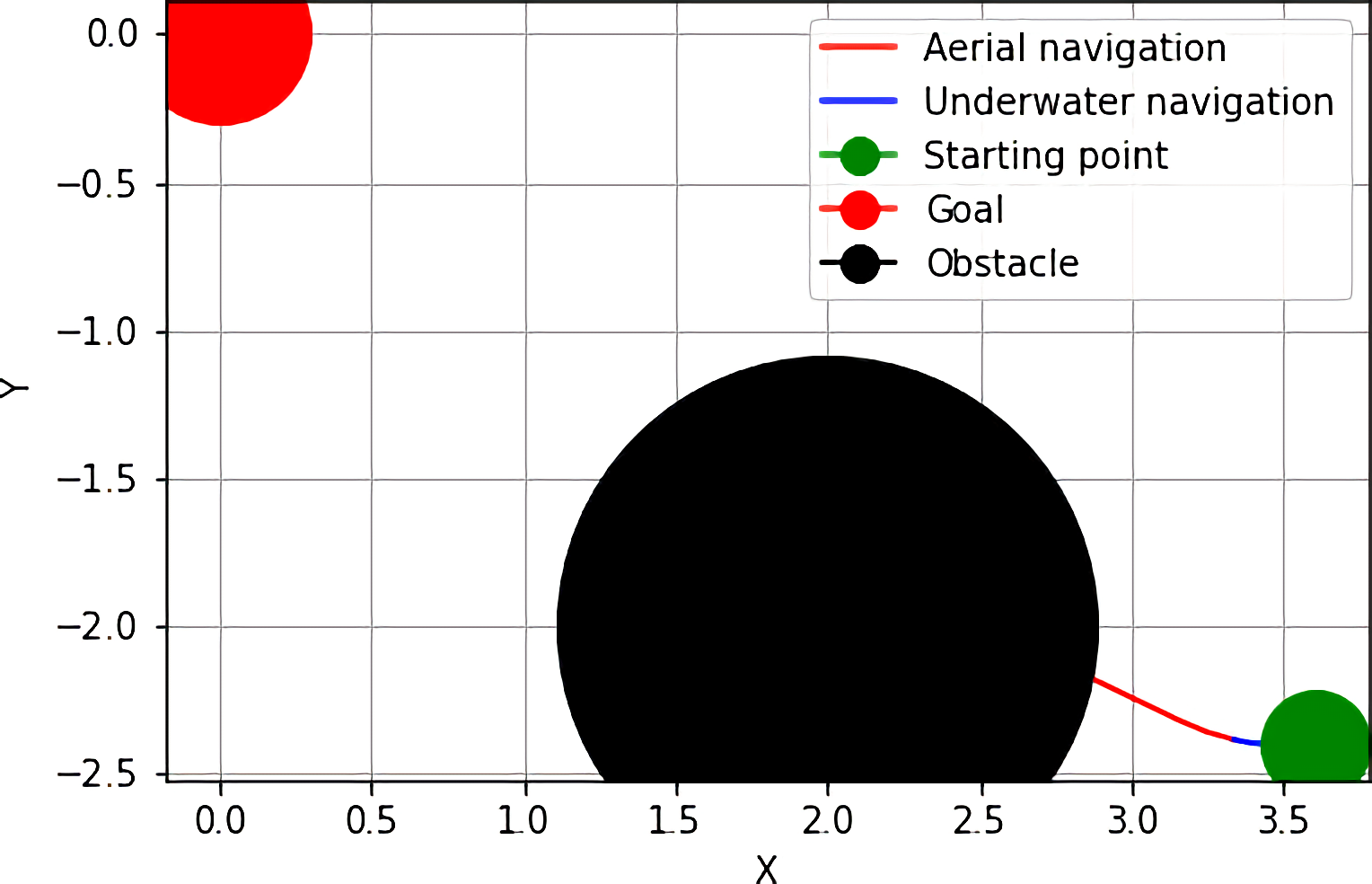}
	\end{minipage}}
 \hfill
 \vspace{-1mm}
\caption{Behavior sample of each approach tested in the second environment.}
\label{fig:traj_nav_2}
\vspace{-6mm}
\end{figure*}

Besides collecting the average time spent in the air and the water and its standard deviation by each one of the 100 tests shown in Table \ref{table:mean_std}, we also tried to analyze the actions generated by each agent in each scenario of the 100 trial through each step of the navigation. We can see these collected results in Fig. \ref{fig:cmd} with the mean action value and its standard deviation at each step. Finally, we collected a navigation sample among each one of the 100 trials and showed the navigation performed in Fig. \ref{fig:traj_nav_1} for the first scenario and in Fig. \ref{fig:traj_nav_2} for the second scenario.

%% file: 5_discussion.tex
\section{Discussion}
\label{discussion}

The extensive validation of our proposed reveals a proper behavior for both agents in both environments. The agents were able to learn environmental difficulties, realize the medium transition, and generalize enough to perform the mapless navigation through air and water. Besides that, our agents also learned to bypass obstacles to arrive at a target, where traditional algorithms fail. We have validated two powerful approaches, which managed to understand the complex behaviors in realistic simulation and that can be used in the real world if guarantying range findings and relative localization data.   

Specifically, we can also observe that our deterministic approach performed considerably better than our 3DNDRL-S approach in the first and simpler scenario with just simulated wind and a water tank. However, in the second scenario with simulated drilling risers, we can observe based on the results of Table \ref{table:mean_std} and Figs. \ref{fig:traj_nav_1} and \ref{fig:traj_nav_2} that the better exploration policy of the 3DNDRL-S approach made it have a similar or even better performance than our deterministic algorithm. Finally, it is interesting to notice that, besides the fact that our two approaches have fundamental differences in its implementation, the actions generated by each of them at each step of testing when averaged 100 times are quite similar, manly in the second and more complex scenario. We can observe this robustness and consistency in Fig. \ref{fig:cmd}.

%% file: 6_conclusion.tex
\section{Conclusions}
\label{conclusion}

In this paper, we have proposed two distinct Deep-RL-based approaches for mapless goal-oriented navigation of a Hybrid Unmanned Aerial Underwater Vehicle (HUAUV) in both, aerial and underwater environments. The methods only use the relative location of the vehicle and range samples. Results obtained in the two trained scenarios show that the robot was capable to perform the complex task of medium transition successfully and learn to bypass objects, like drilling risers, to achieve the desired target.

More specifically, we can examine that with a simple sensory-based structure, both Deep-RL-based approaches had comparable performances. 
However, while they managed to bypass the risers and execute a route in the second environment, the geometric tracking controller leads the robot to collide with them. We can also conclude that Deep-RL approaches are suitable for the development of applications that require continuous control for robots, such as HUAUVs, in complex environments using range sensing systems, and few states. The consistency of the results obtained can also be attributed to a rewarding scheme that is simple and precise, such as those proposed.

As future works, we aim to expand the study with more Deep-RL algorithms and compare them with other traditional algorithms. We are also aiming to test our approaches with our real HUAUV.

%% file: 7_acknowledgment.tex
\section*{Acknowledgment}


This work was partly founded by the CAPES, CNPq and PRH-ANP.

%% file: 8_references.tex
\bibliographystyle{./IEEEtran}
\bibliography{./main}